\documentclass{article}
\usepackage{ijcai18}

\usepackage{times}
\usepackage{xcolor}
\usepackage{soul}
\usepackage[utf8]{inputenc}
\usepackage[small]{caption}
\usepackage{algorithm}  
\usepackage{algorithmic}

\usepackage{graphicx}
\usepackage{subfigure}
\usepackage{multirow}
\usepackage{amsmath,amssymb}
\usepackage[export]{adjustbox}

\title{Anchor-based Nearest Class Mean Loss for Convolutional Neural Networks}

\author{
Fusheng Hao$^{1, 2}$, 
Jun Cheng$^3$, 
Lei Wang$^3$, 
Xinchao Wang$^4$,
Jianzhong Cao$^1$, 
Xiping Hu$^3$, 
Dapeng Tao$^3$
\\ 
$^1$ Xi'an Institute of Optics and Precision Mechanics, CAS, 
$^2$ University of Chinese Academy of Sciences \\
$^3$ Shenzhen Key Laboratory of Virtual Reality and Human Interaction Technology, \\
Shenzhen Institutes of Advanced Technology, CAS\\
$^4$ Department of Computer Science, Stevens Insititute of Technology\\
haofusheng@opt.cn,
\{jun.cheng, lei.wang1, xp.hu\}@siat.ac.cn,
xinchao.wang@stevens.edu \\
cjz@opt.ac.cn, dapeng.tao@gmail.com
}

\begin{document}

\maketitle

\begin{abstract}
Discriminative features are critical for machine learning applications. Most existing deep learning approaches, however, rely on convolutional neural networks (CNNs) for learning features, whose discriminant power is not explicitly enforced. In this paper, we propose a novel approach to train deep CNNs by imposing the intra-class compactness and the inter-class separability, so as to enhance the learned features' discriminant power. To this end, we introduce anchors, which are predefined vectors regarded as the centers for each class and fixed during training. Discriminative features are obtained by constraining the deep CNNs to map training samples to the corresponding anchors as close as possible. We propose two principles to select the anchors, and measure the proximity of two points using the Euclidean and cosine distance metric functions, which results in two novel loss functions. These loss functions require no sample pairs or triplets and can be efficiently optimized by batch stochastic gradient descent. We test the proposed method on three benchmark image classification datasets and demonstrate its promising results.
\end{abstract}

\section{Introduction}
Deep learning methods~\cite{NIPS2012_4824,2014arXiv1409,7298594} have demonstrated superior results in many computer vision tasks, including image recognition even on the very-large-scale benchmark such as ImageNet~\cite{5206848}. Most deep learning methods exploit the strong representation capacity of convolutional neural networks (CNNs) and are often supervised by the softmax loss (SoftMax). This loss function, however, does not explicitly drive CNNs to learn discriminative features.

One recent trend to make better use of the strong representation capacity of CNNs is to explicitly encode the discriminant into the model. For example, Hadsell et al. proposed a Siamese network with a contrastive loss for dimensionality reduction~\cite{1640964}, which learned a mapping such that similar points in the input space were mapped to nearby points in the low dimensional manifold. Schroff et al. extended the contrastive loss to a triplet loss by introducing a third anchor point which was sampled from the training set, and minimized the distance between an anchor and a positive of the same identity and meanwhile maximized the distance between the anchor and a negative of different identity~\cite{7298682}. However, training samples needed to be carefully selected, as the number of training sample grows at the rate of $O(N^2)$, where $N$ is the number of training samples.

\begin{figure*}[!htb]
\centering
\subfigure[Random Deep Features]{
\centering
\includegraphics[height=3.8cm]{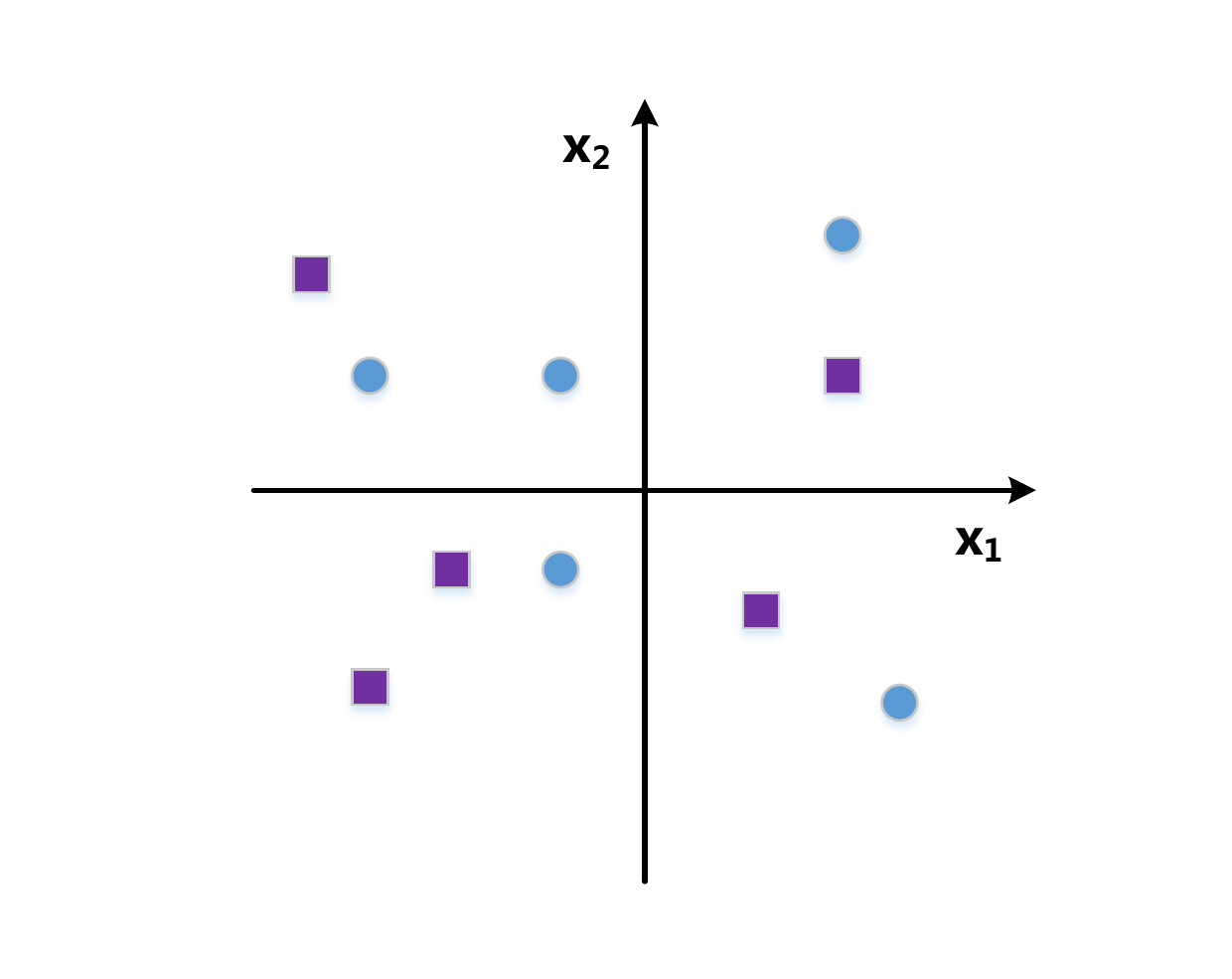}
}%
\hfill
\subfigure[Anchor-based C-NCM]{
\centering
\includegraphics[height=3.8cm]{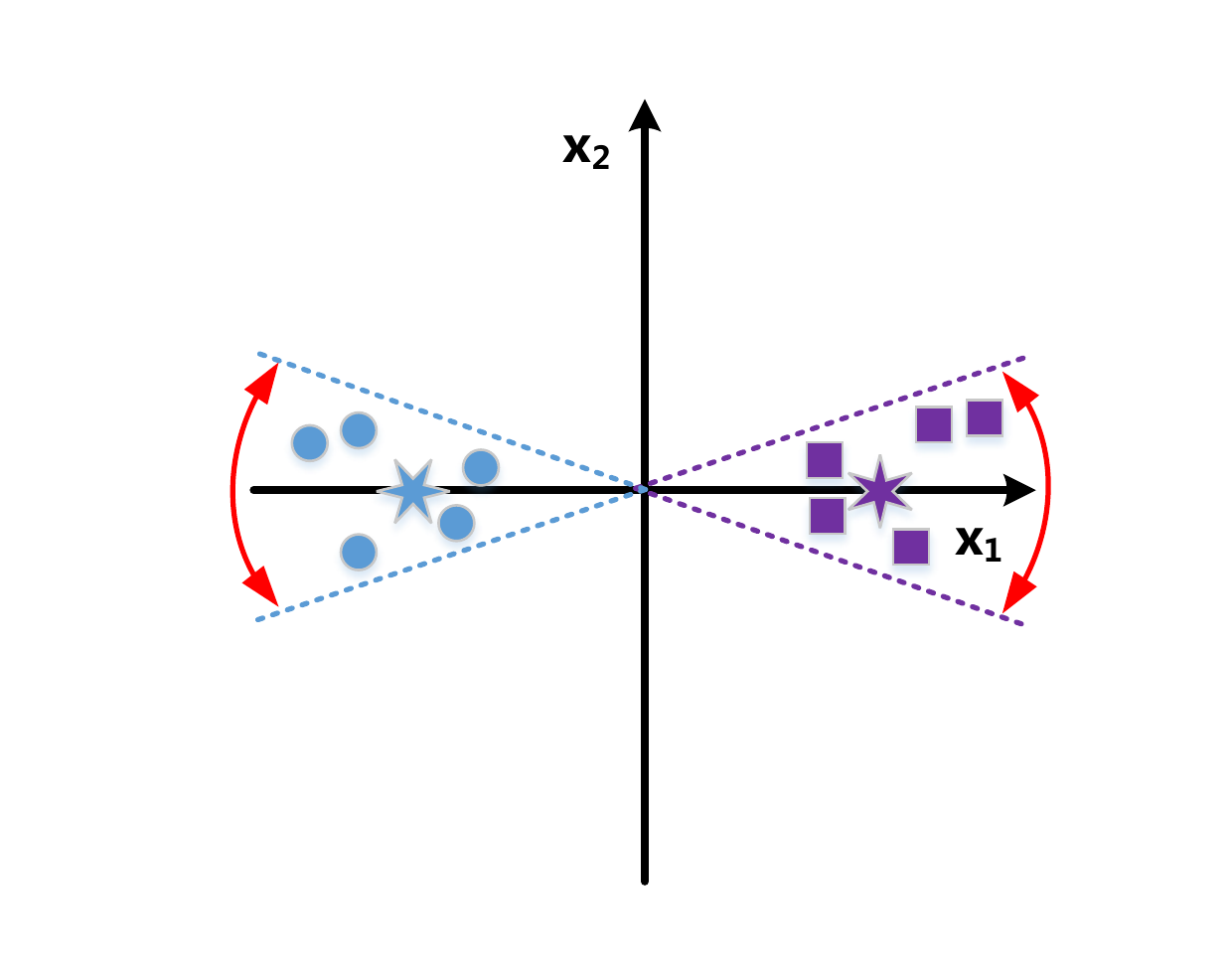}
}%
\hfill
\subfigure[Anchor-based E-NCM]{
\centering
\includegraphics[height=3.8cm]{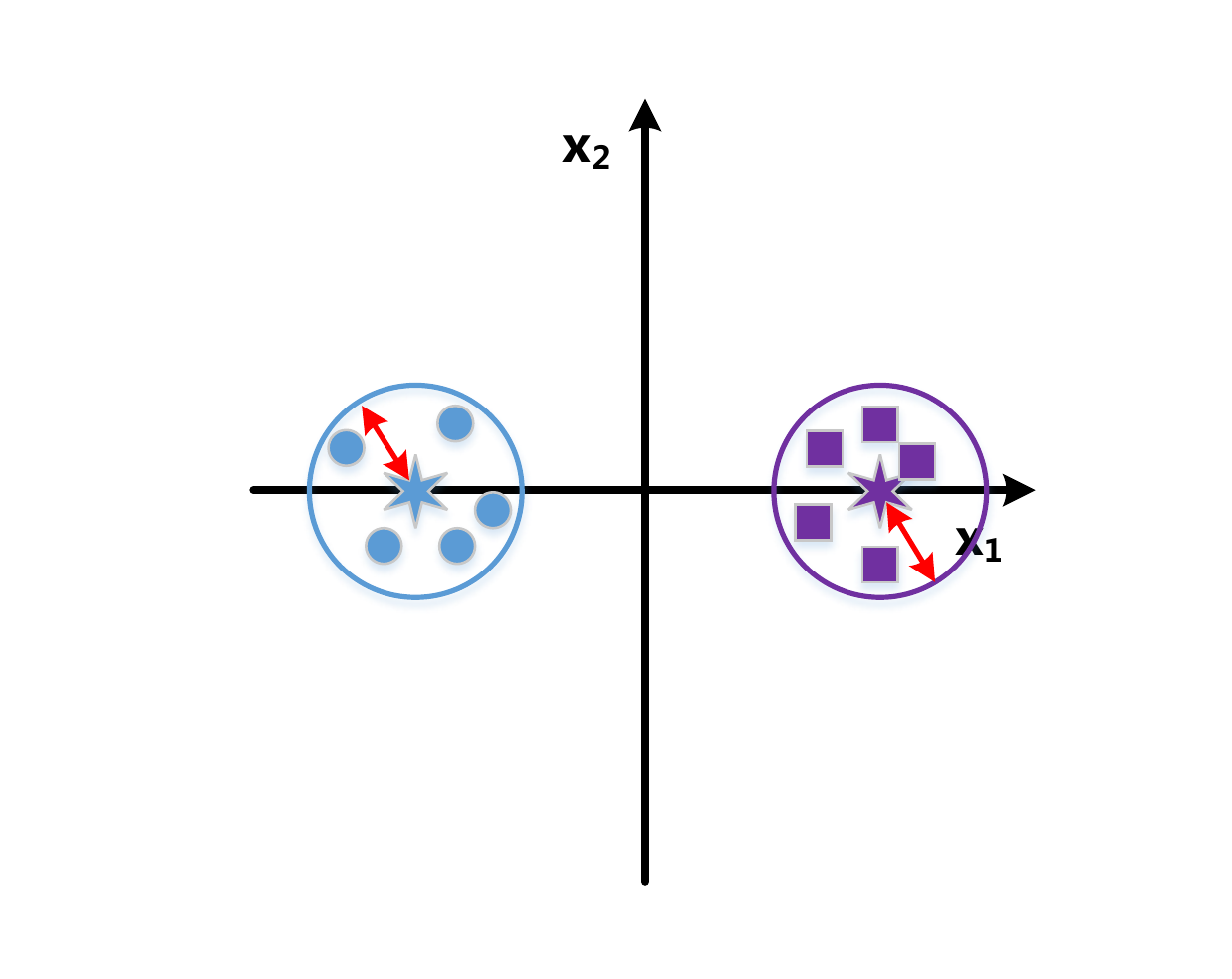}
}%
\vspace{-1mm}
\caption{One toy example of the anchor-based NCM loss. The deep features of the two classes are denoted by purple squares and blue circles respectively. The corresponding anchors for each class are represented by the Hexagonal star of the same color. The deep features are constrained to be as close as possible to the anchors, measured by the cosine or Euclidean distance.}
\vspace{-4mm}
\label{fig:skeleton}%
\end{figure*}

In this paper, we propose a novel approach to train deep CNNs by enforcing unequivocal intra-class compactness and inter-class separability. In contrast to prior approaches, ours explicitly encourages CNNs to learn discriminative features without relying on sample pairs or triplets, and can be efficiently optimized with batch stochastic gradient descent (SGD). This is achieved by introducing to our objective function the notion of anchors, based on which a Nearest Class Mean (NCM) loss is raised. Here, anchors refer to predefined vectors that are fixed during training. The anchors are regarded as class centers, each of which is assigned to a particular class. We foster the learned features to gather around the anchors of the same label, and meanwhile expect the anchors to be as separate as possible. Specifically, we derive anchors using the following two principals:
~~\begin{itemize}
\item The anchors have the unit norm, i.e., they are sampled from a unit hypersphere;
\item The angle between any two anchors is large.
\end{itemize}
Our objective function, which will be discussed in detail in Section 3, involves computing the distance measure between the two points, for which the only requirement is it being differentiable. We consider two common and simple distances, Euclidean and cosine distances, which further leads to two novel loss functions. One toy example is shown in Figure~\ref{fig:skeleton}, highlighting the effect of anchor-based cosine and Euclidean distances on the random deep features. Note that, the proposed method is not limited to the Euclidean or cosine distance and any differentiable distance metric functions can be readily adopted.

The approach of Wen et al.~\cite{Wen2016} also relies on center-based Euclidean distance to learn discriminant features. Unlike our method, it performed updates of class centers based on mini-batch. However, the number of training samples for each class in a training batch is random and thus the centers of any two different class may be very close to each other or have very large difference in norm. These two factors result in large perturbations on the class centers and further lead to slow convergence. Furthermore, it alone cannot learn meaningful features due to the trivial solutions like all deep features being mapped to the origin. By contrast, our method fixes the anchors and avoids such perturbations, yielding discriminative features as well as promising results. Also, our method requires no tunable hyper-parameters, making it easy to be deployed in practice. The approach of Liu et al.~\cite{Liu:2016:LSL:3045390.3045445}, on the other hand, introduces a large-margin softmax loss (L-Softmax), which requires the angle between different classes to exceed a margin. It requires a huge amount of hyper-parameters that are difficult to tune. Also, its backward propagation becomes very demanding when the angle margin is large as the triangle formula must be expanded. Our method however directly constrains the angle between the deep features and anchors, which avoids the tricky strategy of updating the deep features to accelerate convergence. Further, our anchors are selected from the deep feature space to ensure minimal inter-class angular margin between classes.

Our contribution is therefore, introducing the novel anchor-based loss functions into the deep neural works, which does not require sample pairs or triplets and can be efficiently optimized with batch SGD. It is free of tunable parameters and thus very easy to be deployed. Our method leads to very promising results on several standard benchmarks, demonstrating the effectiveness of the proposed method.

\begin{algorithm*}[!htb]   
\caption{The learning algorithm for CNNs with the anchor-based nearest class mean loss}
\label{alg:Framwork}   
\begin{algorithmic}[1]
\REQUIRE ~~\\
Input training images $\{(x_i,y_i)\}_{i=1}^N$ and anchor vectors $\{a_c\}_{c=1}^C$; \\
Initializing the parameters of the network $W$, learning rate $\epsilon$, dropout ratio $p$, max iterations $MaxIter$, batch size $B$, assigning anchor vector $a_c$ to the category $c$ and selecting the distance metric function $M$:  Euclidean or cosine.
\ENSURE Parameters of the network;
\FOR {$iter \in \{1,2,\dots,MaxIter\}$}
\STATE{
Sample a batch from the training set  $\{(x_i,y_i)\}_{i=1}^N$\\
Initialize the gradient $\nabla W = 0$ \\
\FOR {$b \in \{1,2,\dots, B\}$}
\STATE{
Compute $\{M(f_W(x_b),a_{1}), M(f_W(x_b),a_{2}), \dots, M(f_W(x_b),a_{C})\}$ \\
Compute the gradient $\nabla_W L(x_b)$ using (\ref{gradient}) with the parameter settings $\{N=1, x=x_b, y_1=y_b\}$ \\
Accumulate the gradient $\nabla W = \nabla W + \nabla_W L(x_b)$
}
\ENDFOR \\
Update the parameters $W$ with $\nabla W$
}
\ENDFOR
\end{algorithmic}  
\end{algorithm*}

\section{Related Works}
Previous works that focus on encouraging CNNs to learn discriminative features fall into two main categories: the ones relying on distance metric learning and those adopting the class centers. Distance metric learning learns a mapping for the input space of data from a given collection of pair of similar/dissimilar points that preserves the distance relationships in the training data~\cite{DML}. A strong feature representation is critical for distance metric learning performance. Inspired by the success of the contrastive loss~\cite{1640964}, several works have adopted this loss function to address related problems. For example, Sun et al. combined the contrastive loss with softmax to reduce intra-personal variations while enlarging inter-personal differences, achieving impressive face verification accuracy~\cite{DeepID2}. Very deep CNNs result in superb representation capacity, further improving face verification accuracy~\cite{DeepID3}. Li et al. proposed a more general contrastive loss function to obtain strong feature representations for classification~\cite{ijcai2017-308}.

The other category of algorithms adopt the notion of class center, defined as the mean feature vector of its elements. Weinberger et al. proposed a large-margin nearest neighbor approach, its goal being that the k-nearest neighbors always belonged to the same class while examples from different classes were separated by a large margin~\cite{LMNN}. Further, some subsequent works utilized the strong representation capacity of CNNs and introduced loss functions to constrain CNNs to map the deep features from the same identity as closely as possible and from the different identity as far as possible. For example, Liu et al. extended the center loss~\cite{Wen2016} and proposed a congenerous cosine loss, which cooperatively increased similarity within classes and enlarged variation across categories~\cite{2017arXiv170206890L}. Snell et al. adopted the class center concept for few-shot learning~\cite{2017arXiv170305175S}.

Our method differs from the existing methods in that, we introduce the concept of anchors, which are regarded as class centers and fixed during training, and propose two novel loss functions that directly constrain CNNs to map each sample to the corresponding class center as close as possible, where these centers are as isolated as possible.

\section{The Proposed Method}
Firstly, we briefly review the basic nearest class mean classifier and describe its neural version, and then analyze the limitations of the neural version of the nearest class mean classifier and detail our proposed anchor-based nearest class mean loss. Finally, we explore its relationship to existing methods.

\subsection{Nearest Class Mean Classifier}
Given $N$ training images from $C$ classes $\{(x_i,y_i)\}_{i=1}^N$, where $y_i$ is the ground-truth label of image $x_i$, the trivial nearest class mean classifier represents classes by their mean feature vector of its elements. A test image $x$ is assigned to the class $c^{\ast} \in {\{1,\dots,C\}}$ with the closest mean, which can be formulated in the following form:
\begin{align}
  c^{\ast} & = \mathop{\arg\min}_{c \in {\{1,\dots,C\}}}M(x,\mu_c)
\label{ncm}
\end{align}
\begin{align}
  \mu_c = \frac{1}{N_c}\sum_{i:y_i=c}x_i
\label{mu}
\end{align}
where $M$ is the distance metric between the test image $x$ and the class mean $\mu_c$, and $N_c$ is the number of training images in class $c$.

CNNs have a strong feature representation capacity and can be regarded as a feature extractor $f_W:x \to \mathbb{R}^d$, where $W$ is learnable parameters of CNNs and $d$ is the dimension of the output deep features. By replacing the image $x$ or $x_i$ in equations (\ref{ncm}) and (\ref{mu}) with $f_W(x)$ or $f_W(x_i)$, we obtain the neural version of the nearest class mean classifier.

\subsection{Anchor-based Nearest Class Mean Loss}
The standard optimization algorithm for training CNNs is batch SGD or its variants, such as Adam~\cite{adam}. In each iteration, a batch of training images are sampled randomly from the training sets to update the learnable parameters of CNNs. Therefore, the number of training images for a specific category in a batch may be any number from zero to batch size. This raises an issue that how to update the class mean vectors with limited and uncertain number of training samples. We need to carefully design an update rule for the mean vector of each class, especially when $N_c$ ($\forall c \in \{1,\dots,C\}$) is much larger than the batch size.

\begin{figure*}[!htb]
\centering
\subfigure[Training Set (SoftMax)]{
\centering
\includegraphics[height=3.6cm]{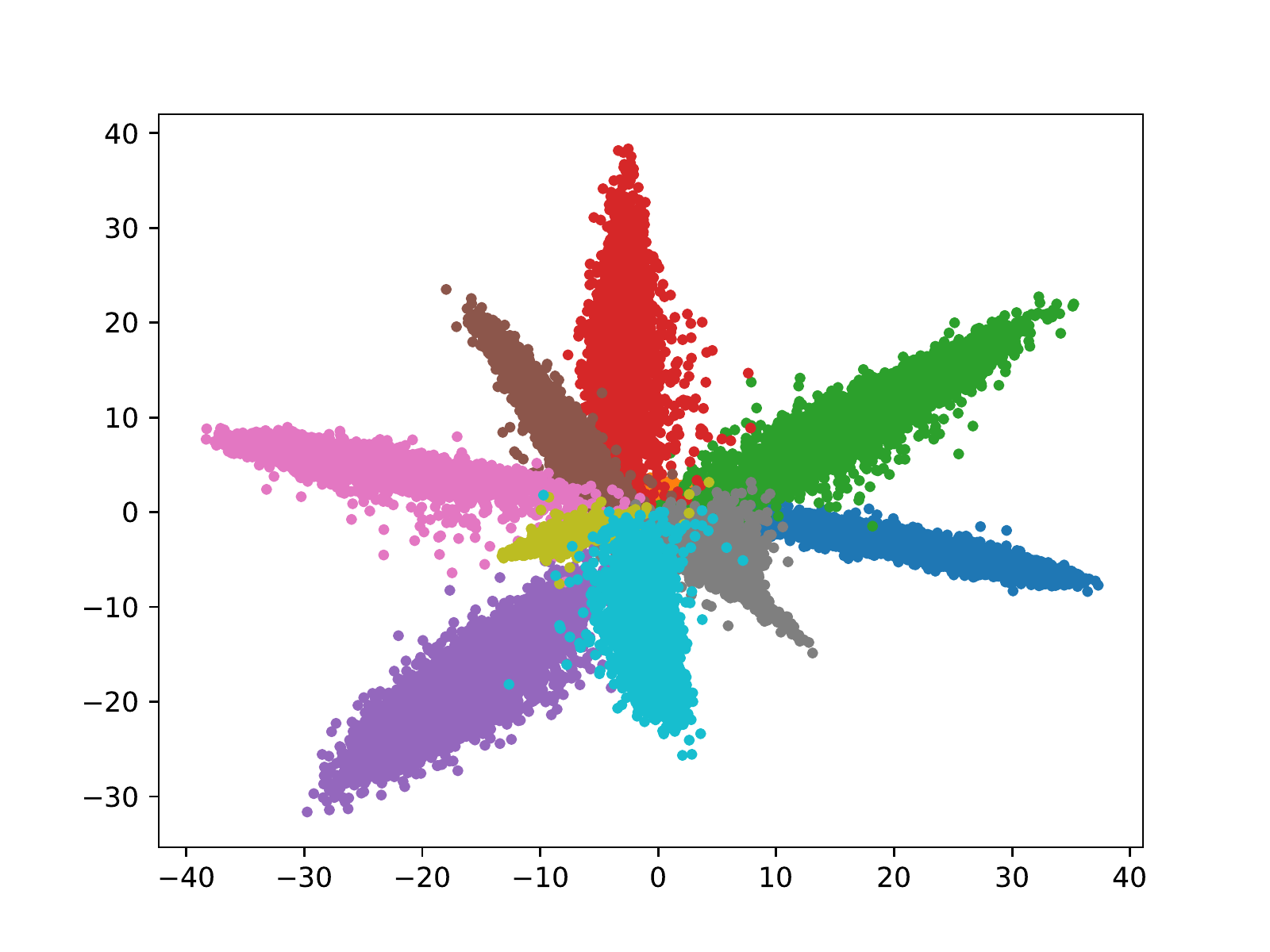}
}%
\hfill
\subfigure[Training Set (C-NCM)]{
\centering
\includegraphics[height=3.6cm]{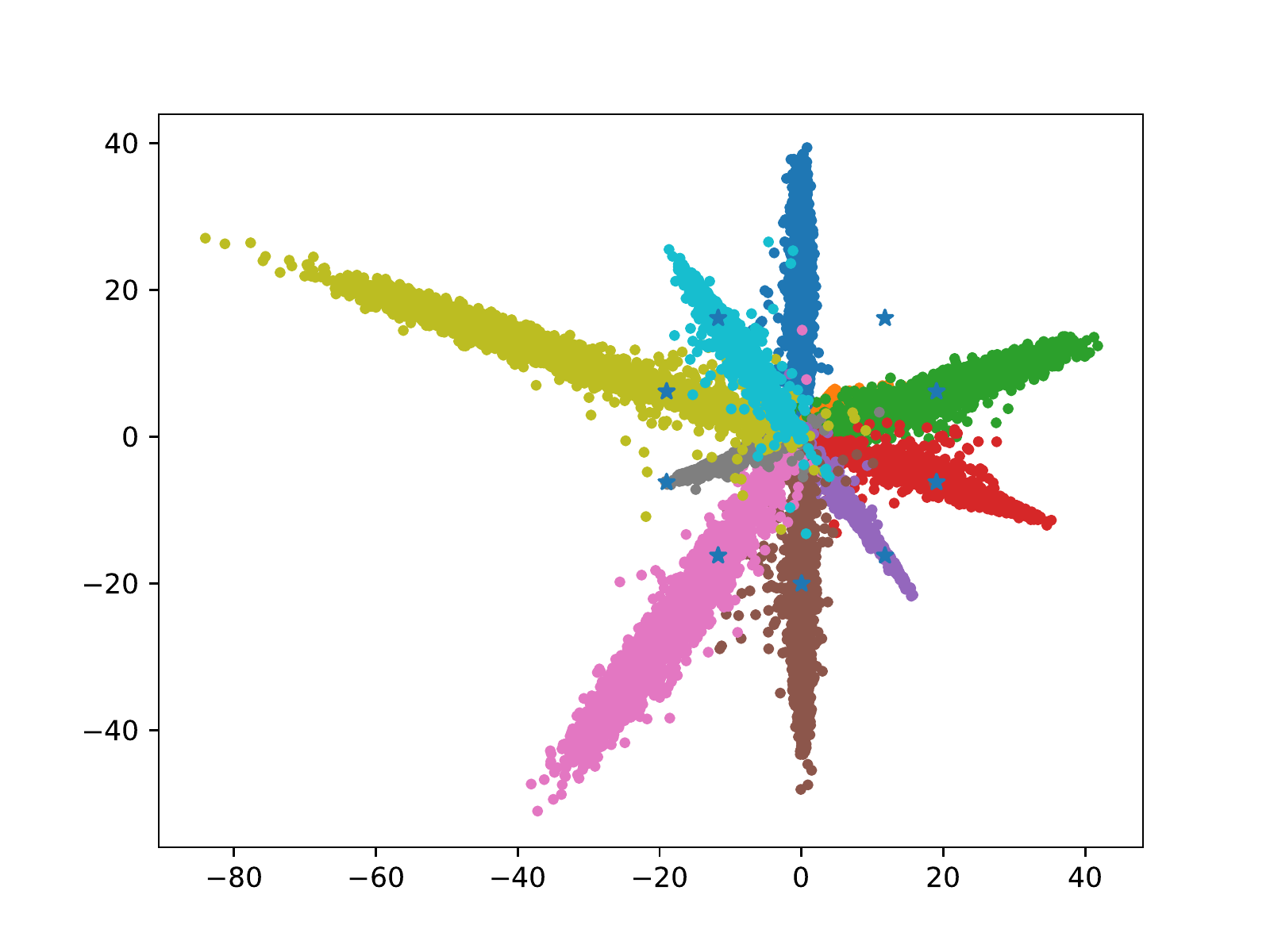}
}%
\hfill
\subfigure[Training Set (E-NCM)]{
\centering
\includegraphics[height=3.6cm]{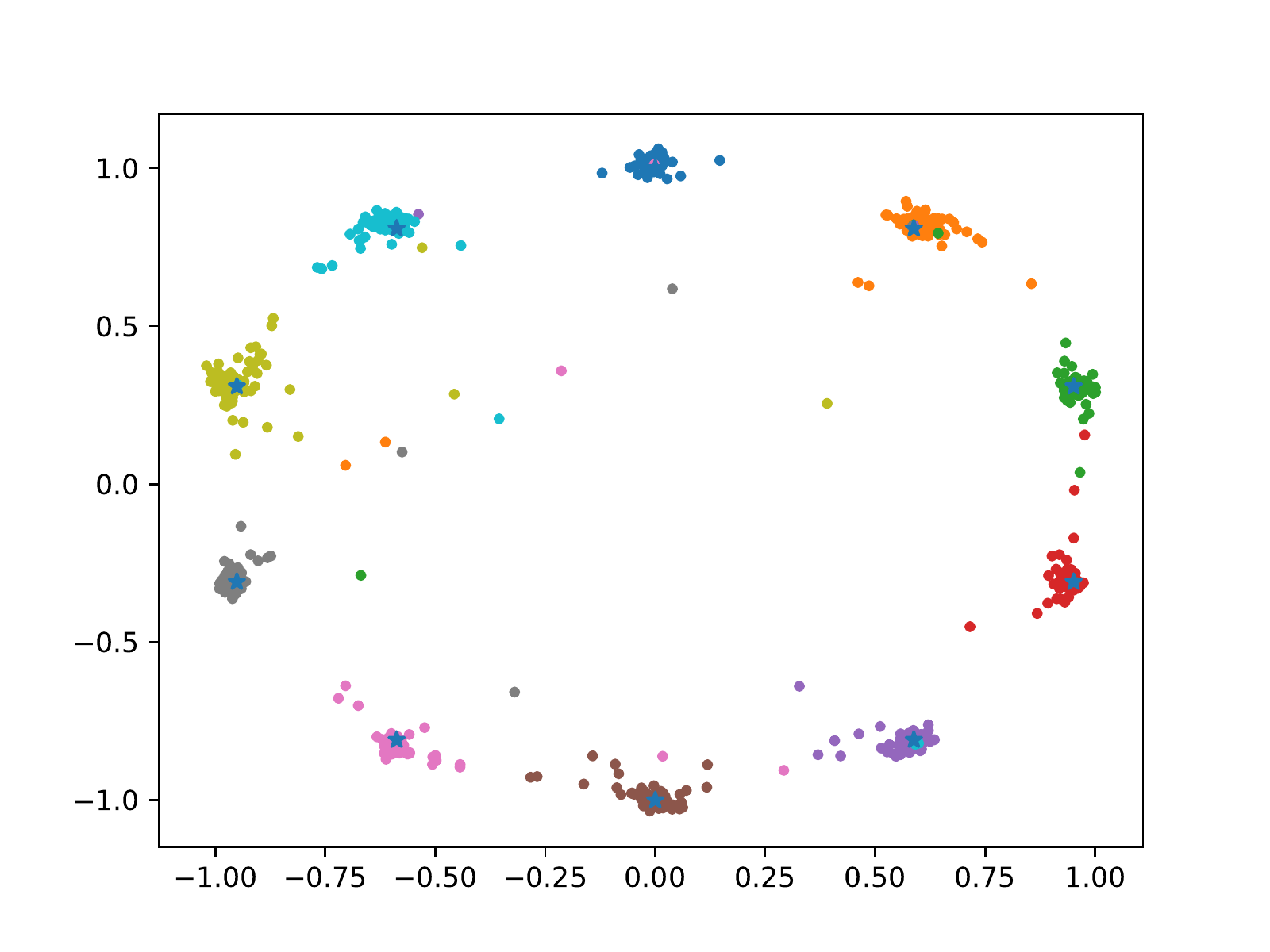}
}%
\\
\subfigure[Test Set (SoftMax)]{
\centering
\includegraphics[height=3.6cm]{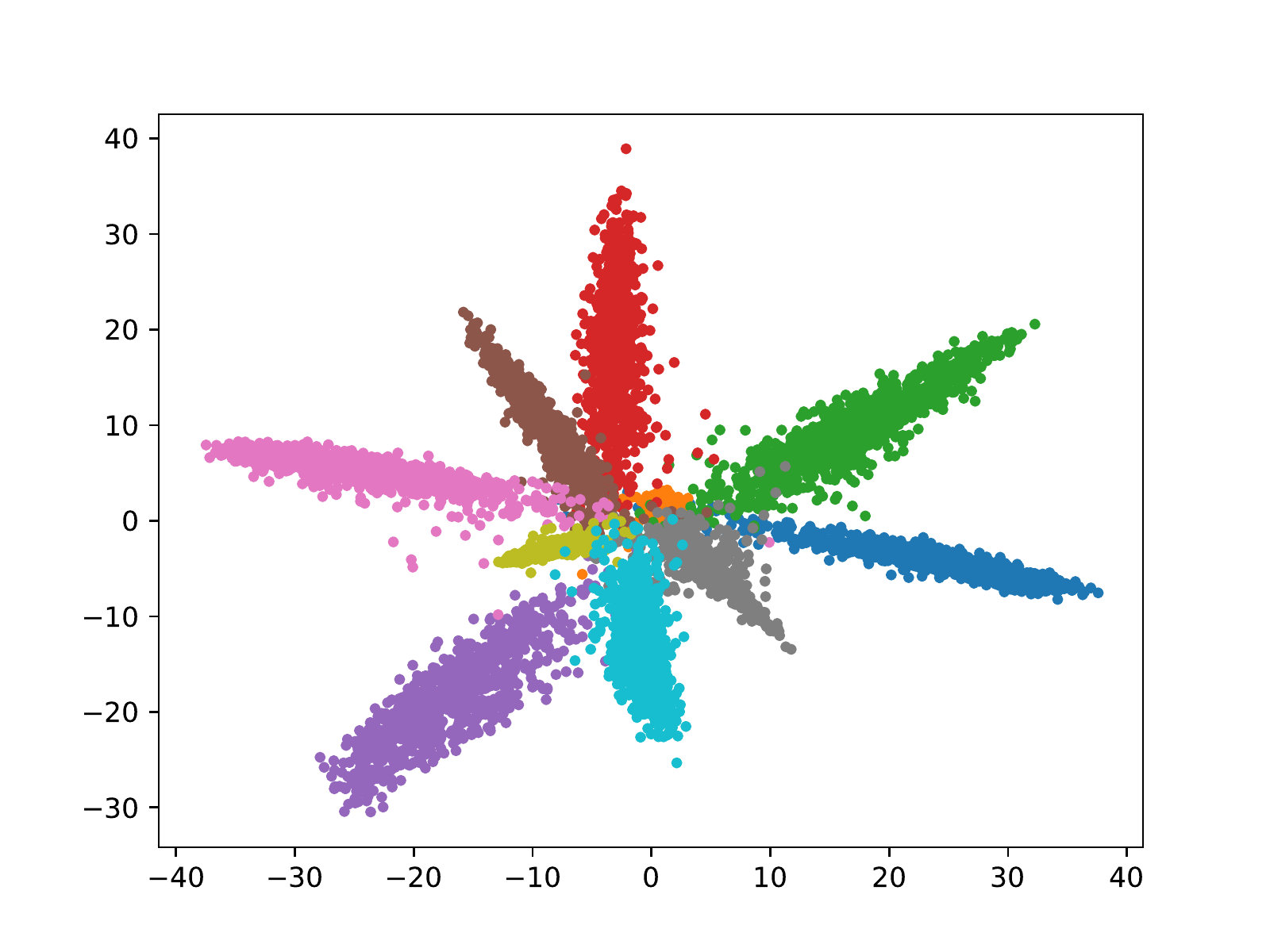}
}%
\hfill
\subfigure[Test Set (C-NCM)]{
\centering
\includegraphics[height=3.6cm]{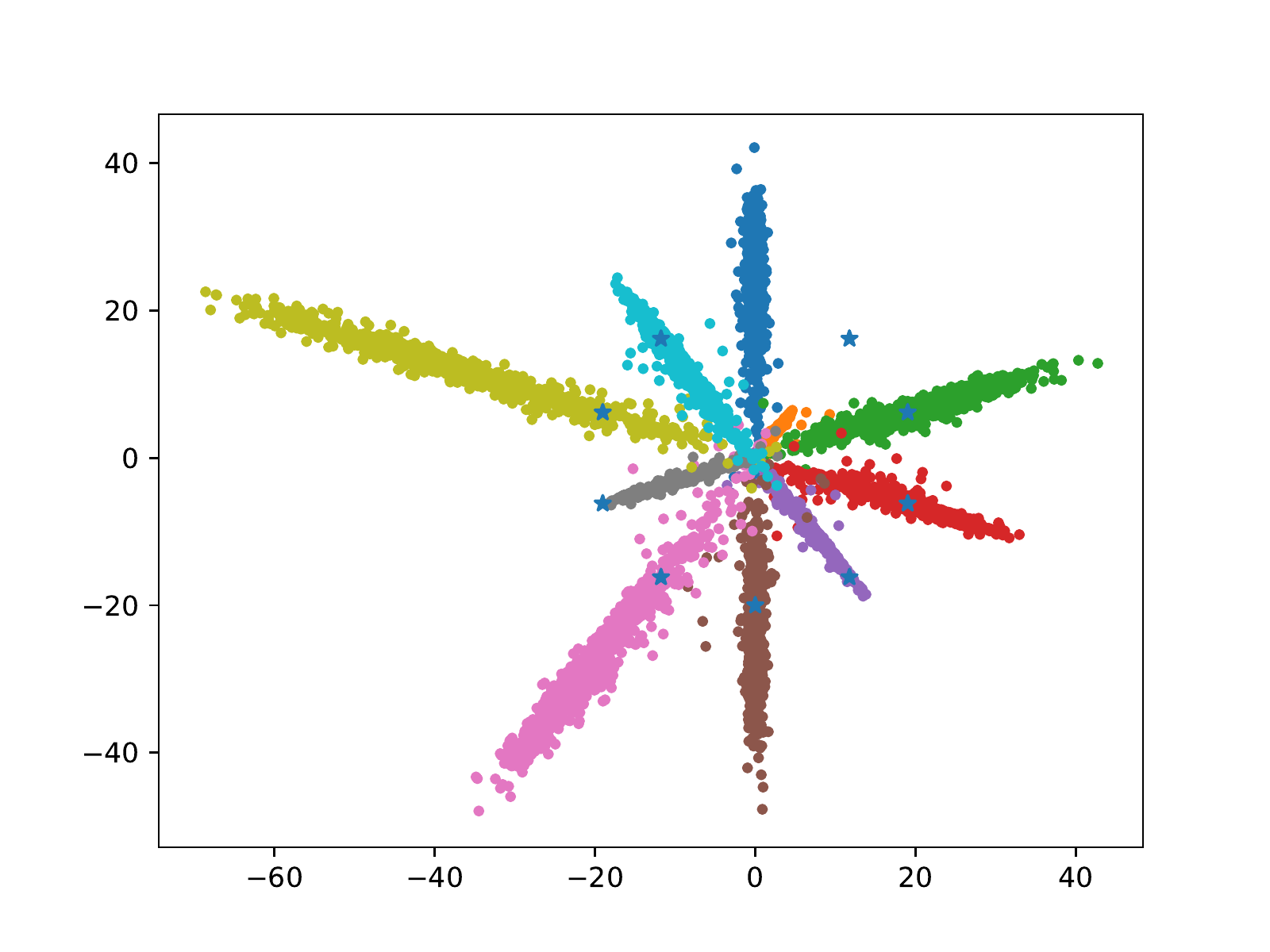}
}%
\hfill
\subfigure[Test Set (E-NCM)]{
\centering
\includegraphics[height=3.6cm]{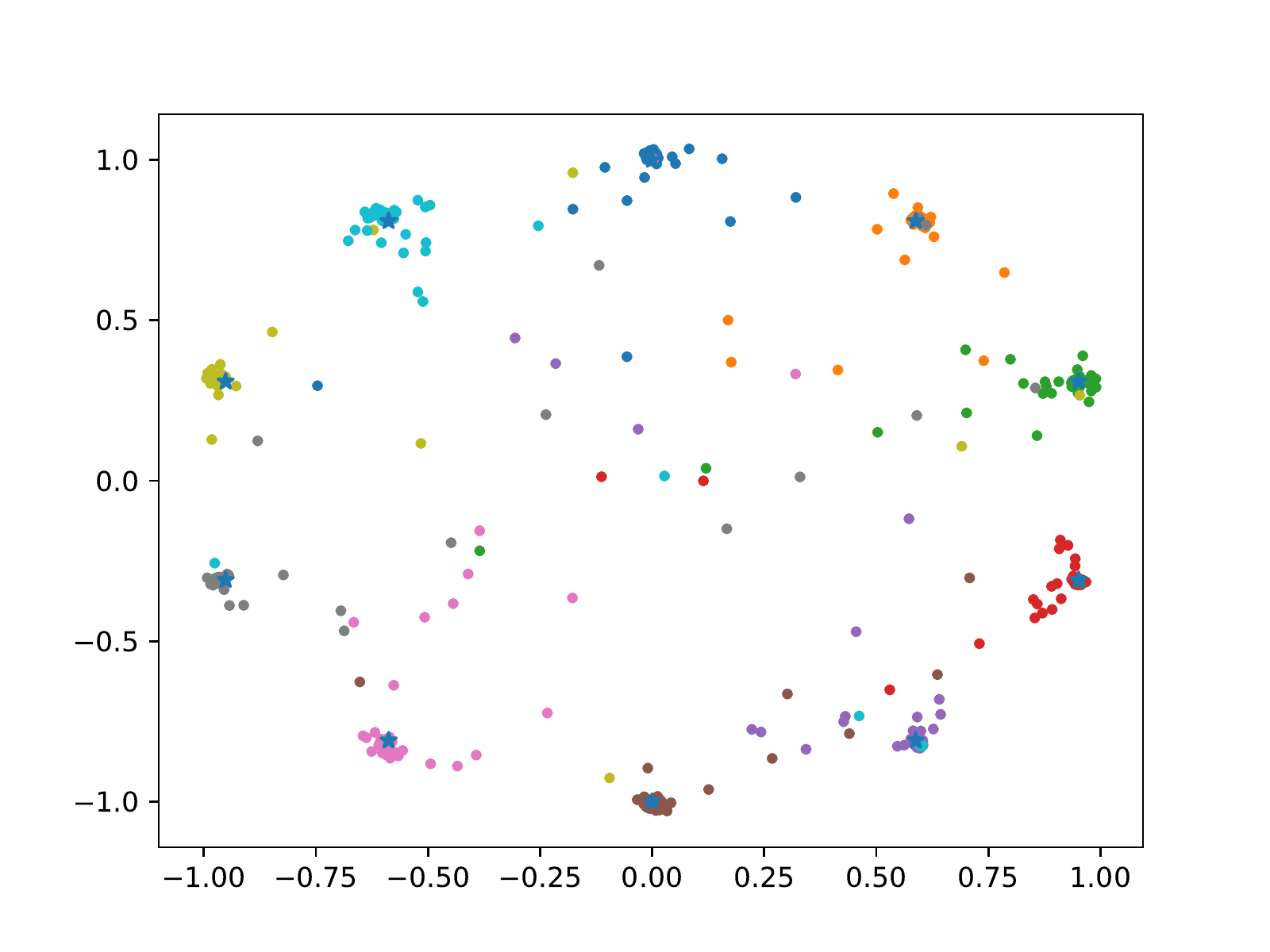}
}%
\vspace{-1mm}
\caption{Deep features visualization of the SoftMax, C-NCM and E-NCM loss functions on the MNIST dataset. The network architecture for producing these deep features is illustrated in Table~\ref{net}. The anchors for C-NCM and E-NCM are indicated by *. Note that the vast majority of deep features for E-NCM are near the corresponding anchors. For better visualization, we have magnified the norm of the anchors for C-NCM 20 times to match the scale of the deep features.}
\label{visualization}%
\vspace{-3mm}
\end{figure*}

In order to avoid a tricky updating rule for the class mean vector, we replace the class mean vector with a set of anchors. These anchors are a set of predefined vectors ${\{a_c\}_{c=1}^C}$ ($a_c \in \mathbb{R}^d$), each of which is randomly assigned to one category and regarded as the class center for that class. An image $x$ is assigned to the class $c^{\ast} \in {\{1,\dots,C\}}$ with the closest mean to the corresponding anchor:
\begin{align}
  c^{\ast} & = \mathop{\arg\min}_{c \in {\{1,\dots,C\}}}M(f_W(x),a_c)
\label{pncm}
\end{align}
Further, to interpret the anchor-based nearest class mean loss from the perspective of probability, we adopt softmax function to represent the categorical distribution of $x$ over $C$ classes. Given an image $x$, the probability for a class $c$ is defined as:
\begin{align}
  p(c \vert x) & = \frac{ exp(-M(f_W(x),a_c)) } { \sum_{c'=1}^C exp(-M(f_W(x),a_{c'}))}
\label{prob}
\end{align}

To train CNNs with the proposed anchor-based nearest class mean loss, we directly minimize the negative log-likelihood of the correct predictions of the training images:
\begin{align}
  L & = -\frac{1}{N} \sum_{i=1}^N \sum_{c=1}^C I(y_i=c) ln(p(y_i \vert x_i))
\label{loss}
\end{align}
where $I(y_i=c)$ is the indicator function that equals one if its arguments is true and zero otherwise. For optimization, we need the derivative of loss function (\ref{loss}) with respect to $W$:
\begin{multline}
  \nabla_W L = -\frac{1}{N} \sum_{i=1}^N \sum_{c=1}^C I(y_i=c) \{ \nabla_W(-M(f_W(x),a_c)) - \\
  \frac{\sum_{c'=1}^C exp(-M(f_W(x),a_{c'})) \nabla_W(-M(f_W(x),a_{c'}))}{\sum_{c'=1}^C exp(-M(f_W(x),a_{c'}))} \}
\label{gradient}
\end{multline}
Considering that CNNs can be viewed as a potent universal function approximator, we choose two simple distance metric functions: Euclidean and cosine, although any differentiable distance metric $M$ is compatible with the anchor-based nearest class mean loss. The concrete forms of the two distance metrics are described as:
\begin{align}
  M(f_1, f_2)_{E} & = \sqrt{(f_1 - f_2)^2}
\label{E}
\end{align}
\begin{align}
  M(f_1, f_2)_{C} & = 1 - \frac{f_1 \cdot f_2}{\left\| {f_1} \right\|_2 \left\| {f_2} \right\|_2}
\label{C}
\end{align}
where, $f_1$ and $f_2$ are the vectors from $\mathbb{R}^d$. They are abbreviated as E-NCM and C-NCM, respectively. The derivatives of the two distance metrics are straightforward.

Given training set $\{(x_i,y_i)\}_{i=1}^N$, CNNs are constrained to map each sample to the corresponding anchor as close as possible. Ideally, all deep features are mapped to the corresponding anchors, which inspires us to design a rational way to select the anchor set to ensure inter-class dispersion and enhance intra-class compactness. To preserve the deep feature balance, we expect the norms of these anchors to be equal. Further, these anchors should be as dispersed as possible to maximize the inter-class dispersion and minimize intra-class compactness of deep features. These two principles can be mathematically formulated as:
\begin{align}
  \lVert a_c \rVert_2=1,\ \forall c \in {\{1,\dots,C\}}
\label{norm}
\end{align}
\vspace{-5mm}
\begin{align}
  \frac{a_c \cdot a_{c'}}{\left\| {a_c} \right\|_2 \left\| {a_{c'}} \right\|_2}\ \geq \cos(\theta_M),\forall c \ne c' 
\label{angle}
\end{align}
where $\theta_M$ is the predefined angle margin. We use polar coordinates to represent points on the hypersphere, and obtain the anchors by meshgriding the angle coordinates and computing the coordinate values. For example, in the two-dimensional space, the points on the unit circle are represented by $(\cos \theta, \sin \theta)$; by meshgriding $2\pi$ with interval $\frac{2\pi}{C}$ and computing the coordinate values, we obtain $C$ anchors in this space. We summarize the steps for our proposed approach in Algorithm 1.

\begin{table*}[!htb]
\caption{Net architectures for different benchmark datasets. Conv0-3.x denotes convolution units that may contain multiple convolution layers. E.g., [$3\times3$, 64]$\times$4 denotes 4 cascaded convolution layers with 64 filters of size $3\times3$. Each convolutional layer is followed by a batch normalization layer and a ReLU layer, which are omitted for brevity.}
\centering
\begin{tabular}{ |c|c|c|c|c| }
\hline
Layer & MNIST (for Figure~\ref{visualization}) & MNIST & CIFAR10/CIFAR10+ & CIFAR100 \\ \hline
\hline
Conv0.x & [$3\times3$, 64]$\times$1 & [$3\times3$, 64]$\times$1 & [$3\times3$, 64]$\times$1 & [$3\times3$, 96]$\times$1 \\ \hline
Conv1.x & [$3\times3$, 64]$\times$3 & [$3\times3$, 64]$\times$3 & [$3\times3$, 64]$\times$4 & [$3\times3$, 96]$\times$4 \\ \hline
Pool1 & \multicolumn{4}{ |c| }{$2\times2$ Max, Stride 2} \\ \hline
Dropout1 & \multicolumn{4}{ |c| }{dropout ratio $p$} \\ \hline
Conv2.x & [$3\times3$, 64]$\times$3 & [$3\times3$, 64]$\times$3 & [$3\times3$, 96]$\times$4 & [$3\times3$, 192]$\times$4 \\ \hline
Pool2 & \multicolumn{4}{ |c| }{$2\times2$ Max, Stride 2} \\ \hline
Dropout2 & \multicolumn{4}{ |c| }{dropout ratio $p$} \\ \hline
Conv3.x & [$3\times3$, 64]$\times$3 & [$3\times3$, 64]$\times$3 & [$3\times3$, 128]$\times$4 & [$3\times3$, 384]$\times$4 \\ \hline
Pool3 & \multicolumn{4}{ |c| }{$2\times2$ Max, Stride 2} \\ \hline
FC & 2 &256 & 256 & 512 \\ \hline
\end{tabular}
\label{net}
\vspace{-4mm}
\end{table*}

\subsection{Relation to Existing Methods}
~~\textbf{Relation to Gaussian mixture model.} If we choose the Euclidean distance metric for anchor-based nearest class mean loss, the definition in (\ref{prob}) may also be interpreted as giving the posterior probabilities of a generative model where $p(c \vert x)=N(x_i;a_c,\Sigma)$ is a Gaussian with mean $a_c$, and covariance matrix $\Sigma$~\cite{6517188}. The class probabilities $p(c)$ are set to be uniform over all classes. Our approach differs significantly from the traditional Gaussian mixture model~\cite{prml} in that the mean vectors are predefined and independent of the actual feature representations. We utilize the strong representation capacity of CNNs and constrain CNNs to simulate the simplified Gaussian mixture model.

\textbf{Relation to SoftMax.} The proposed anchor-based nearest class mean loss differs from SoftMax in two aspects: 1) the deep feature dimensions for anchor-based nearest class mean loss are not limited, while the softmax loss requires the deep feature dimensions to be equal to the class number; 2) with the same dimension, the proposed anchor-based nearest class mean loss constrains CNNs to pull deep features of the same identity to the corresponding anchors, while the softmax loss constrains CNNs to pull the elements of deep features corresponding to classes far greater than the others.

\textbf{Relation to L-Softmax.} L-Softmax is defined as the combination of a softmax function and the last fully connected layer with the requirement that the angle between different classes exceeds a margin. This additional condition complicates the backward propagation of L-Softmax, especially when the angle margin is large. The proposed C-NCM loss function simplifies the backward propagation by pulling the deep features to the fixed anchors. Further, the anchors are selected to ensure minimal inter-class angular margin between classes.

\section{Experiments}
In this section, we present our experimental validations. We first introduce the datasets we used and then show the experiment setup. We finally show comparative results and provide discussions.

\subsection{Datasets}
The proposed E-NCM and C-NCM loss functions are evaluated in the context of the image classification task. We conduct experiment using three standard benchmark datasets: MNIST~\cite{Lecun1998}, CIFAR10~\cite{cifar}, and CIFAR100~\cite{cifar}. MNIST consists of a training set of 60000 $28\times28$ gray-scale handwritten digits of 10 classes and a test set of 10000 samples. CIFAR10 consists of 60000 $32\times32$ color images from 10 classes, of which 50000 are used for training and 10000 for test. CIFAR100 consists of 60000 $32\times32$ color images from 100 classes. Each class of CIFAR100 has only 500 images for training and 100 images for test. If not specified, no data augmentation is applied and the only preprocessing is a global normalization to zero mean and unit variance. In addition, we also test the performance of both loss functions on the CIFAR10 with data augmented (CIFAR10+). We follow the standard data augmentation in~\cite{Liu:2016:LSL:3045390.3045445} for training: 4 pixels are padded on each side, and a $32\times32$ crop is randomly sampled from the padded image or its horizontal flip.

\subsection{Experimental Setup}
~~\textbf{Net architectures.} SoftMax is one of the most commonly used loss function for training CNNs, which is regarded as the baseline in all experiments. We also compare the proposed E-NCM and C-NCM loss functions with Hinge Loss and L-Softmax. For fair comparisons, we use the net architectures as described in~\cite{Liu:2016:LSL:3045390.3045445}. Considering the strong representation capacity of CNNs, the deep features may be pulled to the corresponding anchors easily. Therefore, we also introduce dropout layers in the net architectures to alleviate the issue of overfitting. Specifically, dropout layer is added after the first two max pooling layers. For convolution layers, the kernel size is $3\times3$ and 1 padding (if not specified) to keep the feature map size unchanged. The details of the net architecture for each dataset are described in Table~\ref{net}.

\textbf{Parameter settings.}
We implement E-NCM and C-NCM with the PyTorch deep learning framework. By stacking on the top of CNNs, the two proposed loss functions constrain CNNs to map each training sample to the corresponding anchor. We adopt the standard batch SGD for optimization. Following the settings in L-Softmax, we use a weight decay of 0.0005 and momentum of 0.9 and the batch size is 256. As we pull deep features of the same identity to the fixed anchor, more training iterations may be helpful to further improve the performance. To this end, we start with a learning rate of 0.1, divide it by 10 when the training loss plateaus. Our model requires approximately 90 epochs for the optimal performance. In addition, we extensively evaluate the effect of the dropout ratio (with $p=0$, $p=0.1$, $p=0.25$ and $p=0.5$) on the performance of the proposed E-NCM and C-NCM loss functions. If not specified, we take 10 or 100 standard orthogonal bases from the deep feature space.

\begin{table*}[!htb]
\caption{Recognition error rate (\%) on MNIST, CIFAR10/CIFAR10+ and CIFAR100 datasets. + denotes the performance with data augmentation.}
\vspace{-3mm}
\centering
\begin{tabular}{ |c|c|c|c|c| }
\hline
Method & MNIST & CIFAR10 & CIFAR10+& CIFAR100 \\ \hline
\hline
Hinge Loss~\cite{Liu:2016:LSL:3045390.3045445} & 0.47 & 9.91 & 6.96 & 32.90\\ \hline
L-softmax (m=2)~\cite{Liu:2016:LSL:3045390.3045445} & 0.32 & 7.73 & 6.01 & 29.95\\ \hline
L-softmax (m=3)~\cite{Liu:2016:LSL:3045390.3045445}  & 0.31 & 7.66 & 5.94 & 29.87\\ \hline
L-softmax (m=4)~\cite{Liu:2016:LSL:3045390.3045445}  & 0.31 & \textbf{7.58} & 5.92 & 29.53\\ \hline
\hline
C-NCM (p=0.1) & \textbf{0.25} & 8.78 & 5.98 & 30.86\\ \hline
E-NCM (p=0.1) & 0.26 & 8.89 & \textbf{5.67} & \textbf{28.14}\\ \hline
\end{tabular}
\label{results}
\vspace{-4mm}
\end{table*}

\subsection{Results}
\begin{figure}[!htb]
\centering
\includegraphics[height=3.8cm]{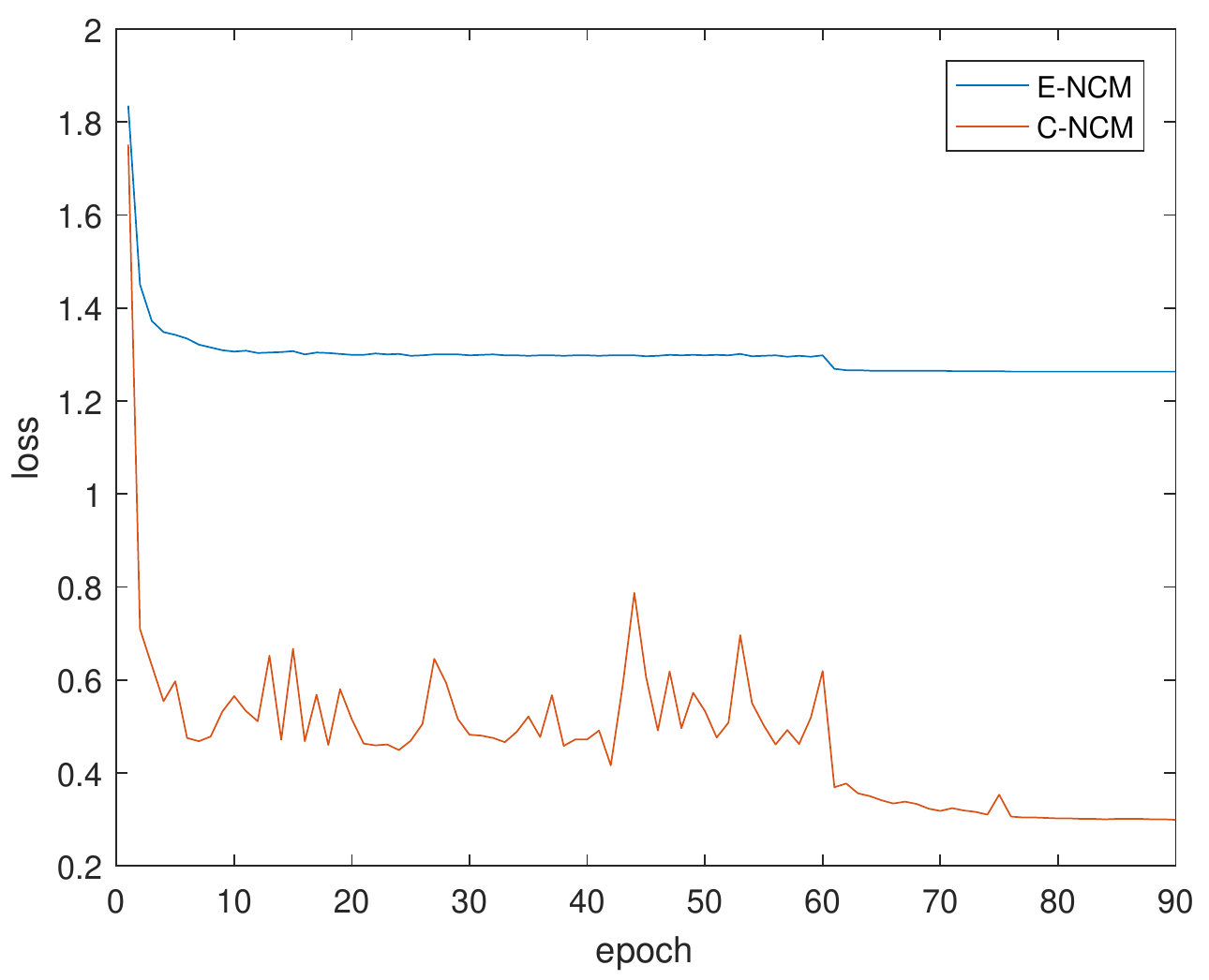}
\vspace{-3mm}
\caption{The convergence of E-NCM and C-NCM on the MNIST dataset.}
\label{convergence}%
\vspace{-4mm}
\end{figure}

~~\textbf{Visualization on MNIST.}
For the convenience of visualization, we set the dimension of the output deep features to 2. As SoftMax requires the dimension of input features to be equal to the number of classes, another fully connected layer with 10 outputs is added after the deep features. The proposed two loss functions are not dependent on the input features dimension. The ten anchors are sampled from the unit circle with equal angle between adjacent anchors. The net architecture for this experiment is illustrated in Table~\ref{net}. The test accuracy for SoftMax, C-NCM and E-NCM are 99.32\%, 99.00\% and 99.53\% respectively. In Figure~\ref{visualization}, we can observe that C-NCM constrains the training samples tighter than SoftMax, which demonstrates the feasibility of pulling deep features to the predefined anchors. The vast majority of training samples are mapped close to the anchors by E-NCM. It is not straightforward to notice that, as most of the sample points reside at the same location. There are a very small amount of outliers in the test phase, some of which are far from the corresponding anchors in both E-NCM and C-NCM. One possible factor is that the training samples for a particular class cannot cover the whole space of the particular class. Based on this observation, we introduce dropout layers in the net architectures to expand sample space in the following experiments. Note that the scales of deep features between different loss functions are different, indicating the necessity of limiting the anchors to be of the same length. In addition, almost all the deep features on MNIST supervised by only the center loss~\cite{Wen2016} is near the origin, which indicates the infeasibility of only the center loss to learn discriminative deep features.
We also show the convergence of E-NCM and C-NCM in~Figure~\ref{convergence}, where both converge within 10 epochs.

\textbf{Ablation study.}
We constrain CNNs to map limited training samples to the corresponding anchors as close as possible, which may result in overfitting and weak generalization. Thus, we introduce data augmentation and dropout to alleviate this issue. Without data augmentation and dropout, C-NCM and E-NCM already achieve better performance than the baseline SoftMax on CIFAR10 and CIFAR100 as shown in Table~\ref{dropout}. The proposed two loss functions benefit even more from the data augmentation and dropout than SoftMax. E-NCM achieves slightly better performance than C-NCM on CIFAR10 and CIFAR10+, but roughly 2\% lower error rate than C-NCM on CIFAR100. Note that, E-NCM not only enforces the angle between training samples and the corresponding anchors as small as possible, but also restricts the norm of deep features to be equal to the anchors. This observation may imply that with limited training samples, stricter constraints could lead to better performance.

\begin{table}[!htb]
\centering
\caption{The effect of dropout layers and data augmentaion on the performance of SoftMax, C-NCM and E-NCM. Recognition error rate (\%) is reported on CIFAR10, CIFAR10+ and CIFAR100 datasets. + denotes the performance with data augmentation.}
\label{dropout}
\vspace{-1mm}
\begin{tabular}{|p{1.4cm}|p{1.2cm}|l|l|l|l|}
\hline
\multicolumn{1}{|c|}{\multirow{2}{*}{Dataset}} & \multicolumn{1}{c|}{\multirow{2}{*}{Method}} & \multicolumn{4}{c|}{Dropout ratio}                                                                               \\ \cline{3-6}
\multicolumn{1}{|c|}{}  & \multicolumn{1}{c|}{}  & \multicolumn{1}{c|}{p=0} & \multicolumn{1}{c|}{p=0.1} & \multicolumn{1}{c|}{p=0.25} & \multicolumn{1}{c|}{p=0.5} \\ \hline
\hline
\multirow{3}{*}{CIFAR10}   &   SoftMax    &   10.50             &   10.22              &   9.81                  &  10.13   \\ \cline{2-6} 
                                            &   C-NCM    &   9.98                 &  \textbf{8.78}    &   8.97                 &   \textbf{8.52}  \\ \cline{2-6} 
                                            &    E-NCM   &   \textbf{9.94}    &  8.89                 &    \textbf{8.43}   &   8.94   \\ \hline
\hline
\multirow{3}{*}{CIFAR10+}   &  SoftMax     &  6.60                &  6.61                 &    6.75                &  7.12   \\ \cline{2-6} 
                                              &   C-NCM    &    6.10                &   5.98                &    6.04                &   6.46  \\ \cline{2-6} 
                                              &   E-NCM    &   \textbf{5.90}    &   \textbf{5.67}   &    \textbf{5.77}   &    \textbf{6.32}  \\ \hline
\hline
\multirow{3}{*}{CIFAR100}   &   SoftMax    &   35.37    &  37.26    &  37.26     &   40.25  \\ \cline{2-6} 
                                             &   C-NCM    &    31.92   &   30.86   &   30.57    &   31.31  \\ \cline{2-6} 
                                             &   E-NCM    &   \textbf{29.41}    &   \textbf{28.14}   &    \textbf{28.11}   &    \textbf{28.68}  \\ \hline
\end{tabular}
\vspace{-4mm}
\end{table}

\textbf{Comparison with the-state-of-the-art.}
We also compare the proposed two loss functions with the-state-of-the-art method L-Softmax. C-NCM explicitly increases the angle margin between different categories, which shares the same goal with L-Softmax. E-NCM imposes stronger constraints on CNNs than C-NCM does. The E-NCM loss function enforces CNNs to map training samples to fall on the corresponding anchors exactly. C-NCM achieves an improvement of 0.06\% over L-Softmax on MNIST as shown in Table~\ref{results}. Both C-NCM and E-NCM achieve worse performance than L-Softmax on CIFAR10. However, with data augmentation and dropout, E-NCM achieves the best performance with an improvement of 0.25\% over L-Softmax while the C-NCM achieves the comparable performance with L-Softmax. The proposed E-NCM loss function achieves the lowest error rate of 28.14\%, which is 1.39\% lower than that of L-Softmax on CIFAR100. For L-Softmax, the hyper-parameters, including base, $\gamma$ and $\lambda$ need to be set empirically. We also conduct extensive experiments using L-Softmax under many parameter configurations with the net architectures in Table~\ref{net}. With dropout layers in the architectures, we do not get better results than the ones reported in the original paper. We therefore take directly the results reported in the paper. Note that, the proposed two loss functions have no hyper-parameters to be specified in advance.

\section{Conclusions}
In this paper, we introduce the notion of anchors, which are predefined vectors and fixed during training, into the objective function of deep neural networks. We propose a novel way to utilize the strong representation capability of CNNs, which is to constrain CNNs to map training samples to corresponding anchors as close as possible, and meanwhile isolate the anchors as much as possible. We apply two commonly-used distance metric functions to measure the distance between deep feature of a training sample and the corresponding anchor, which leads to two novel loss functions: E-NCM and C-NCM. Both loss functions require no hyper-parameters and are easy to be implemented. Extensive experiments are conducted on three benchmark datasets. The experimental results demonstrate the feasibility of enforcing CNNs to map training samples to the corresponding anchors, as well as the effectiveness of the proposed loss functions.

\newpage
\bibliographystyle{named}
\bibliography{ijcai18}

\end{document}